\documentclass{article}

\usepackage[preprint]{neurips_2022}

\usepackage[utf8]{inputenc} 
\usepackage[T1]{fontenc}    
\usepackage{hyperref}       
\usepackage{url}            
\usepackage{booktabs}       
\usepackage{amsfonts}       
\usepackage{nicefrac}       
\usepackage{microtype}      
\usepackage{xcolor}         

\usepackage{graphicx}

\title{The impact of individual information exchange strategies on the distribution of social wealth}

\author{
Yang Shao\\
Hitachi Ltd.\\
1 Chome-280\\
Higashikoigakubo\\
Kokubunji, Tokyo
\and
Hirokazu Atsumori\\
Hitachi Ltd.\\
1 Chome-280\\
Higashikoigakubo\\
Kokubunji, Tokyo
\and
Tadayuki Matsumura\\
Hitachi Ltd.\\
1 Chome-280\\
Higashikoigakubo\\
Kokubunji, Tokyo
\and
Kanako Esaki\\
Hitachi Ltd.\\
1 Chome-280\\
Higashikoigakubo\\
Kokubunji, Tokyo
\and
\\
Shunsuke Minusa\\
Hitachi Ltd.\\
1 Chome-280\\
Higashikoigakubo\\
Kokubunji, Tokyo
\and
\\
Hiroyuki Mizuno\\
Hitachi Ltd.\\
1 Chome-280\\
Higashikoigakubo\\
Kokubunji, Tokyo
}

\begin{document}

\maketitle

\begin{abstract}
Wealth distribution is a complex and critical aspect of any society. Information exchange is considered to have played a role in shaping wealth distribution patterns, but the specific dynamic mechanism is still unclear. In this research, we used simulation-based methods to investigate the impact of different modes of information exchange on wealth distribution. We compared different combinations of information exchange strategies and moving strategies, analyzed their impact on wealth distribution using classic wealth distribution indicators such as the Gini coefficient. Our findings suggest that information exchange strategies have significant impact on wealth distribution and that promoting more equitable access to information and resources is crucial in building a just and equitable society for all.
\end{abstract}

\section{Introduction}

Wealth distribution is a crucial aspect of any society, and understanding how it forms is a complex and multifaceted issue. The importance of studying social wealth distribution lies not only in the fact that wealth is a crucial social resource, but also in the observation that statistical patterns similar to those found in wealth distribution can be identified in many other fields, such as the citation counts of research papers and the level of attention received by celebrities \citep{vermeulen2018fat}. To study the distribution of wealth is essentially to study the distribution of a group of similar social successes represented by the distribution of wealth. One important factor to be considered in wealth distribution is the role of information exchange \citep{coelho2005family, hu2007simulating}. Information exchange plays a role in shaping wealth distribution patterns by allowing individuals and groups to acquire knowledge and resources that are necessary for building wealth. In recent years, advances in technology have made information exchange more accessible and efficient than ever before. The internet and social media platforms, in particular, have transformed the way people communicate and share information. These developments have created new opportunities for people to access information and resources and have made it easier for individuals and groups to collaborate and work together towards common goals. However, not all individuals and groups have equal access to information and resources \citep{nishi2015inequality}. In many societies, there are significant disparities in access to education, technology, and other resources that are necessary for participating in information exchange. These disparities can create barriers that prevent some individuals and groups from fully participating in the economy and building wealth. Moreover, information exchange can also have unintended consequences that can exacerbate existing inequalities. For example, individuals and groups that are already wealthy may have more resources to invest in acquiring and sharing information, which can further consolidate their wealth and power. Despite these challenges, information exchange remains a crucial factor in shaping wealth distribution patterns. In this report, we will explore the ways in which information exchange affects wealth distribution, and the various ways in which society can work to promote more equitable access to information and resources. By understanding the role of information exchange in wealth distribution, we can work towards building a more just and equitable society for all. 
In this research, we use simulation based method to focus on the investigation of the impact of various modes of information exchange on wealth distribution.  

\section{Related works}

Research methods for wealth distribution and information exchange mainly include theoretical analysis, empirical research, and numerical simulation. Theoretical analysis and empirical research mainly focus on the construction of theoretical models and analysis of empirical data, while numerical simulation can more intuitively demonstrate the impact of different information exchange strategies on wealth distribution. Therefore, in this research, we use numerical simulation to construct different information exchange strategies and analyze their impact on wealth distribution. In these numerical simulations, we use some classic models, such as the famous Barabási-Albert Model \citep{albert2002statistical} and Small-World Network Model \citep{watts1998collective}, and make some improvements and extensions based on actual conditions. In each model, we adopt different information exchange strategies and compare their performance in wealth distribution. In addition, in this research, we also use some classic wealth distribution indicators, such as the Gini coefficient \citep{dorfman1979formula} and Lorenz curve \citep{gastwirth1971general}, to measure the distribution of wealth under different information exchange strategies. At the same time, we also consider some possible influencing factors, such as the initial distribution of node wealth and network structure, to more comprehensively evaluate the impact of different information exchange strategies on wealth distribution. 

\subsubsubsection{Basic Principles of Wealth Distribution} 

Wealth distribution is an important field of study in economics and sociology. Economists mainly focus on the disparity of wealth and inequality in the distribution of wealth, while sociologists are more concerned with issues of distribution fairness and social justice. Early research mainly focused on individual economic behavior and the role of market mechanisms, such as Adam Smith's theory of the "invisible hand" and David Ricardo's labor theory of value. However, in recent years, people have increasingly recognized the important role of social networks and information exchange in wealth distribution \citep{nishi2015inequality}. 

\subsubsubsection{Social Networks and Wealth Distribution} 

Social network theory holds that social networks have a significant impact on economic activity and the formation of wealth distribution. In these networks, behaviors such as information exchange, resource sharing, and cooperative transactions between individuals can promote the flow and distribution of wealth. For example, research by Albert-László Barabási and Réka Albert \citep{barabasi1999emergence} shows that social network structure plays an important role in the stability and equilibrium of wealth distribution. At the same time, numerous empirical studies have shown that behaviors such as information exchange and resource sharing between nodes in social networks are crucial for the flow and distribution of wealth.  

\subsubsubsection{Information Exchange and Wealth Distribution}

Information exchange is a crucial factor in wealth distribution \citep{peress2004wealth}. Information asymmetry often leads to inequality in resource allocation and wealth distribution. For example, in financial markets, those with more information can typically earn higher returns, exacerbating wealth inequality. However, transparency and fairness in information exchange can also promote balanced and stable wealth distribution. In recent years, with the continuous development and application of information technology, various new information exchange strategies and platforms have emerged, providing new opportunities to promote the flow and distribution of wealth.  

\subsubsubsection{Simulation Based Research}

With the development of modern economics, more and more researchers begin to pay attention to the use of simulation methods to study economic phenomena \citep{herz1998experiential, moiseev2017agent}. The simulation method is a computer program-based simulation technique that helps researchers conduct experiments and analyzes in a virtual economic environment to explore possible economic changes and policy impacts. Compared with traditional statistical and empirical research methods, simulation methods have stronger theoretical and experimental control capabilities. They can provide researchers with a more realistic picture of the economic environment, allowing them to better simulate the effects of economic changes and policy implementation. In addition, simulation methods can help researchers better understand the nature of economic phenomena and provide more empirical evidence for decision-making. Simulation methods are widely used in economics. For example, they can be used to study the impact of tax policies on different groups \citep{altig2001simulating} or to explore the stability of market regulatory mechanisms \citep{teufel2013review}. 
\citet{pluchino2018talent} used a simulation-based method to show that in Western culture, the dominant elite paradigm, which is characterized by high competition, overlooks the influence of external factors in personal success stories. Success often depends not only on personal qualities, such as talent and intelligence, but also on random factors. The research suggests that it is not reasonable to allocate too much honor or resources to lucky people, and recommends policy measures to improve elite management, diversity of thought, and innovation. However, their research assumes that everyone is a completely independent individual and does not take into account the influence of social networks on opportunity creation (reflected in luck). Their research also did not take into account that in different social cultures, changes in social networks will bring different luck distributions to individuals in different positions. Referring to the aforementioned influence of social network and information exchange on wealth distribution \citep{nishi2015inequality, barabasi1999emergence, peress2004wealth}, we add the influence of social network and information exchange strategy to the simulation, so as to observe and analyze the influence of different information exchange strategies of individuals and different information exchange cultures of groups on the wealth distribution.  

\section{Simulation Methodology}

\subsection{Simulation Setups} 

\begin{figure}
\centering
\includegraphics[scale=0.4]{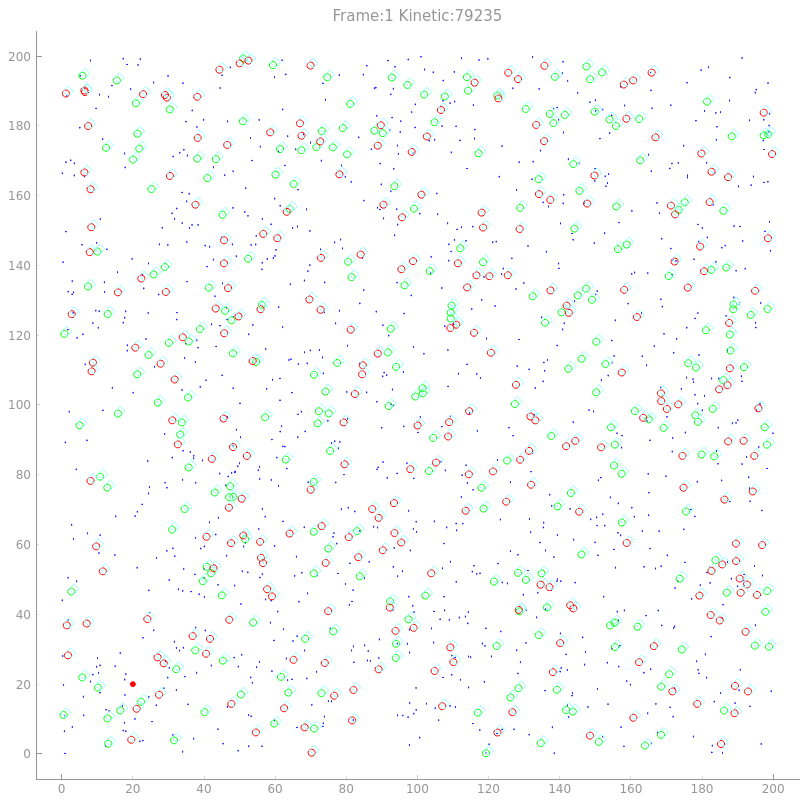}
\caption{An example of initial setup for simulations ($N=1000$ individuals (agents), with different degrees of talent (intelligence, skills, etc.), are randomly located in their positions within a square world of $200*200$ patches with periodic boundary conditions. During each simulation, which covers several dozens of years, they are exposed to a certain number $N_E$ of lucky (green circles) and unlucky (red circles) events, which move across the world following random trajectories (random walks). In this example, $N_E=500$.)}
\label{fig:simulation}
\end{figure}

Our simulations use the TvL model \citep{pluchino2018talent}, an agent model based on a small number of simple assumptions that aims to describe the career evolution of a group of people under the influence of lucky or unlucky random events. We consider $N$ individuals (denoted by blue dots in Fig.\ref{fig:simulation}), each of whom has a talent value of $T_i$ (intelligence, skill, ability, etc.), which obeys a normal distribution around a given mean $m_T$, with a standard deviation of $\sigma_T$, randomly distributed in a square world. The world has a periodic boundary condition (i.e., has a ring topology) surrounded by a certain number of "motion" events (denoted by green and red circles in Fig.\ref{fig:simulation}), some of which are lucky and others unlucky (neutrals are not considered in the model events because they have no significant impact on the individual's life). In Fig.\ref{fig:simulation}, we represent these events as colored circles: lucky events are green and represent a relative percentage of the total number of events, $pL$, and unlucky events are red and represent a percentage of the total number of events $100-pL$. The total number of events $N_E$ is uniformly distributed, but the distribution tends to be completely uniform only when $N_E$ is infinity. In our simulation, $N_E$ is proportional to $N/2$. Thus, at the start of each simulation, different regions of the world randomly distribute more lucky or unlucky events, while other regions are more neutral. Moving circles further randomly within the square lattice (i.e., the world) does not change the fundamental feature of the model, which is that different individuals, regardless of their talents, face different numbers of lucky or unlucky events during their lifetime.  

For a single simulation run, consider a 50-year working life period (from 20 to 70 years) with time steps of half a year for a total of 100 steps. At the start of the simulation, all agents have the same amount of capital of the order $C_i(0) = C(0) (i = 1, ..., N)$, denoting their starting wealth level. The purpose of this selection is to provide no initial advantage to anyone. While the talent of an agent is time independent, the capital of an agent varies over time. During the temporal evolution of the model, i.e., during the lifetime of the agent under consideration, all event circles are randomly moved around the world, and in doing so may intersect with some agent's positions. In each time step, each event circle is moved by a distance of length $v$ in a random direction. The radius of event circles is $r$. When the event circle and the individual change from disjoint to intersecting, we say that the event happened to this agent. After the intersection, the event circle will not disappear.  

According to this, at a given time step (i.e., every half year), there are 3 different possible actions for a given agent $A_i$: 

1. There is no event circle intercepting the location of agent $A_i$: this means that no important events have occurred in the past half a year. Agent $A_i$ does nothing. 

2. The lucky event intercepts the position of agent $A_i$: this means that a lucky event has occurred in the past six months (note that the generation of innovative ideas is considered to be a lucky event that occurs in the brain of the agent). Therefore, agent $A_i$ will increase its capital by an order of magnitude with probability proportional to its talent $T_i$. Only when $rand [0,1] < T_i$, i.e., when the agent is smart enough to benefit from its luck, its capital order will become $C_i(t) = C_i(t-1) + dC$. Here, $dC$ is the average impact of each event on the magnitude of wealth. 

3. Unfortunate event intercepts the position of agent $A_i$: this means that an unfortunate event has occurred within the past half year; thus, agent $A_i$ will reduce its capital by an order of magnitude, i.e., $C_i(t) = C_i(t-1) - dC$. 

The above rules (including changing capital by orders of magnitude in the event of misfortune or luck, the probability of change being proportional to the talent of the agent, etc.) are simple and widely agreed because they are based on commonsense evidence that wealth in every person's life is usually characterized by very rapid growth or decline. In addition, these rules give highly talented people a significant advantage, because they can better exploit the opportunities that luck brings (including the ability to use the good ideas born in their brains). On the other hand, a car accident or sudden illness, for example, is always an unfortunate event, and individual talents play no role in avoiding such events. Thus, we can more effectively generalize the concept of "talent" as "any personal quality that increases the chances of being seized." In other words, the term "talent" broadly refers to intelligence, skill, cleverness, tenacity, determination, hard work, risk-taking, etc.  

Research by \citet{pluchino2018talent} showed that having the advantage of great talent is a necessary but not sufficient condition for attaining very high wealth. However, the matter of building a social network and obtaining information from it can neither be included in individual talent nor simply abstracted into luck. Because of changes in social networks themselves and obtaining information in social networks, both of these things depend not only on the information exchange strategies of individuals, but also on the information exchange strategies of the other party (or more generally, on the information exchange strategies of the group culture). This is a process of mutual influence, and this process cannot be decoupled into two independent variables of talent and luck. So, in order to study the influence of social network and information exchange on wealth distribution, we added the following settings.  

1. The action of agent exchanging information in social network is added. 

2. The action that the agent moves in the space according to the obtained information is added. 

3. The action of the agent to update its social network based on the information obtained is added. 

Therefore, compared with the original research \citep{pluchino2018talent}, our research adds a social network through which agents can share their own information and get information from other agents, so that they can move to richer places (places with more lucky events). Each agent has the following 6 attributes: geographic location, talent value, wealth magnitude, social network links, social network update strategy, and geographic location movement strategy. Among them, the latter 3 items are newly added in our research.  

\subsection{Experiment Designs}

As a preliminary research, we limit the update rules of social networks to the following 4 types: 

1. Random: There is a random social relationship between agents 

2. Location: There is a social relationship between agents whose distance between geographic locations is within a radius $R$.  

3. Wealth: There is a social relationship between agents whose distance between wealth magnitudes is within $nC$ times $dC$.  

4. Talent: There is a social relationship between agents whose distance between talent value is within $nT$ times standard deviation of talent value $T$.  

We limit mobile strategies to the following 3 types: 

1. Random: Random movement 

2. Highest: Follow the agent with the highest order of wealth in the social network 

3. Average: Follow the weighted average position of the agent position in the social network according to the order of wealth 

We use mean wealth, wealth variance, and Gini coefficient 3 indicators to assess the macro-level results of the simulations.  

\section{Results and Analysis}

\begin{figure}
\centering
\includegraphics[scale=0.3]{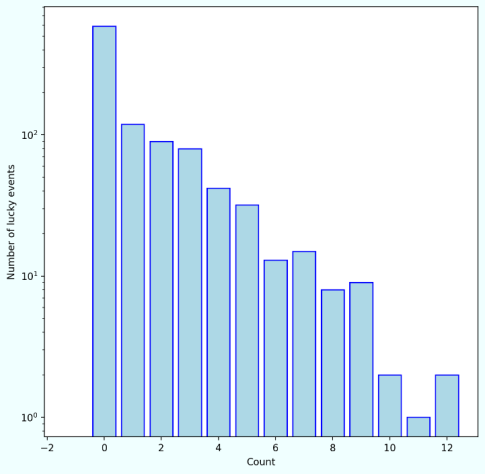}
\includegraphics[scale=0.3]{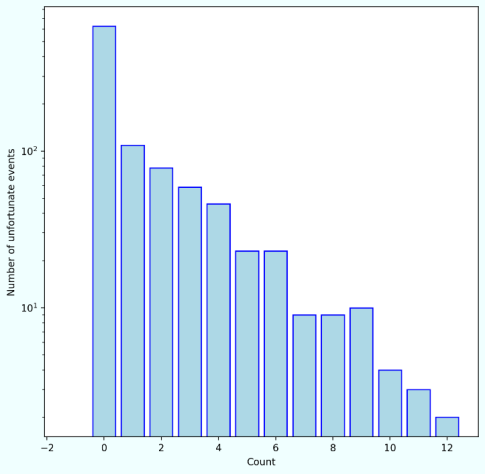}
\caption{The frequency distributions of numbers of lucky (left) and unlucky (right) events are reported
separately on a log-linear scale. It can be seen that both distributions are well fitted by an exponential
distribution with similar negative exponents. (Random-Random)}
\label{fig:rr12}
\end{figure}
\begin{figure}
\centering
\includegraphics[scale=0.3]{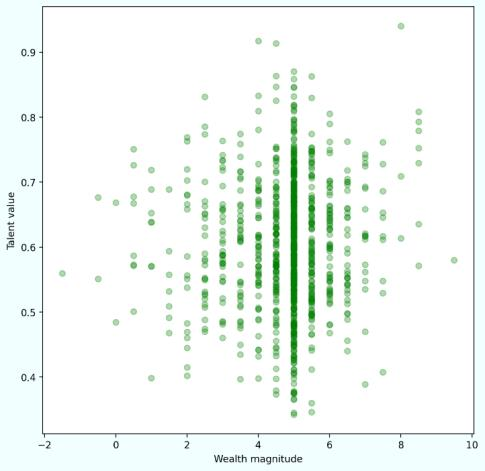}
\includegraphics[scale=0.3]{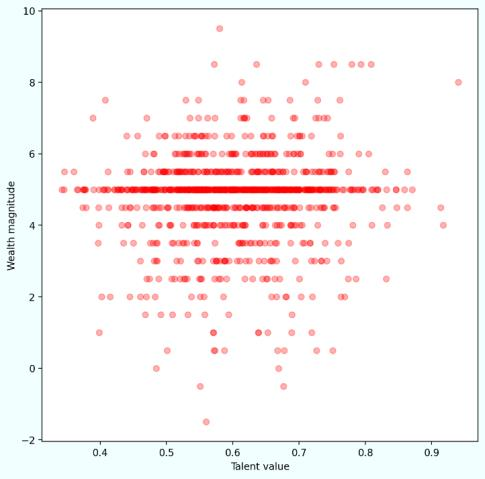}
\caption{In the left figure, talent is plotted as a function of wealth magnitude: the wealthiest people are not
the most talented. In the right figure, wealth magnitude is plotted as a function of talent: the wealthiest agents have talents only around average, while the most talented only have assets around their starting assets.
(Random-Random)}
\label{fig:rr34}
\end{figure}
\begin{figure}
\centering
\includegraphics[scale=0.3]{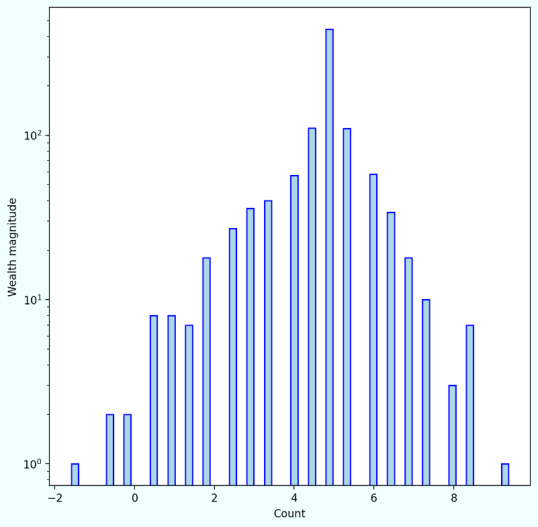}
\caption{The final distribution of wealth among the population (log-lin scale). Despite the normal
distribution of talent, the distribution of wealth shows a strong double-exponential effect. (Random-Random)}
\label{fig:rr5}
\end{figure}

\begin{figure}
\centering
\includegraphics[scale=0.4]{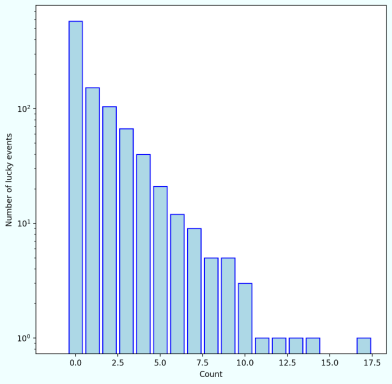}
\includegraphics[scale=0.4]{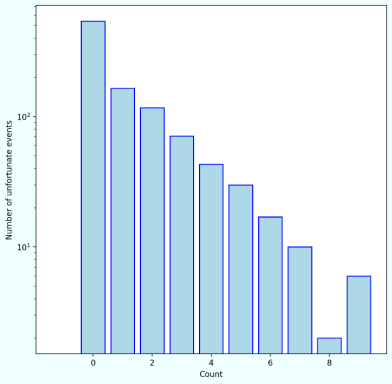}
\caption{The frequency distributions of numbers of lucky and unlucky events (Random-Highest)}
\label{fig:rh12}
\end{figure}

\begin{figure}
\centering
\includegraphics[scale=0.4]{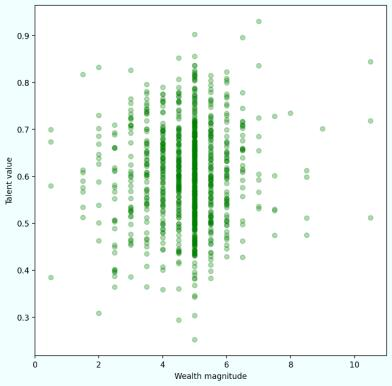}
\caption{Talent-Wealth plots (Random-Highest)}
\label{fig:rh34}
\end{figure}

\begin{figure}
\centering
\includegraphics[scale=0.4]{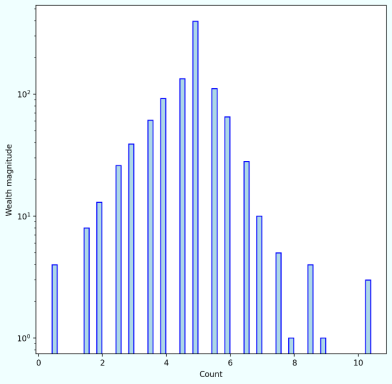}
\caption{The final distribution of wealth among the population (Random-Highest)}
\label{fig:rh5}
\end{figure}

\begin{figure}
\centering
\includegraphics[scale=0.4]{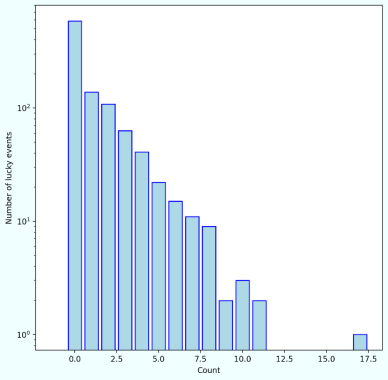}
\includegraphics[scale=0.4]{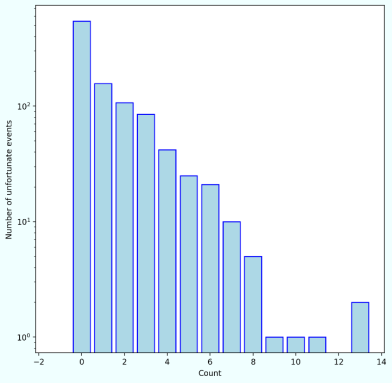}
\caption{The frequency distributions of numbers of lucky and unlucky events (Random-Average)}
\label{fig:ra12}
\end{figure}

\begin{figure}
\centering
\includegraphics[scale=0.4]{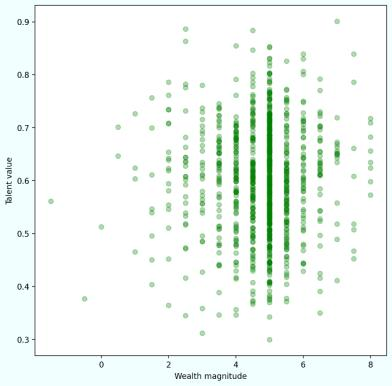}
\caption{Talent-Wealth plots (Random-Average)}
\label{fig:ra34}
\end{figure}

\begin{figure}
\centering
\includegraphics[scale=0.4]{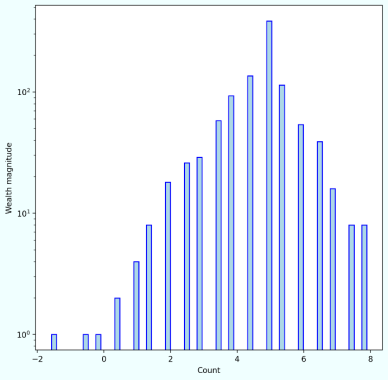}
\caption{The final distribution of wealth among the population (Random-Average)}
\label{fig:ra5}
\end{figure}

\begin{figure}
\centering
\includegraphics[scale=0.4]{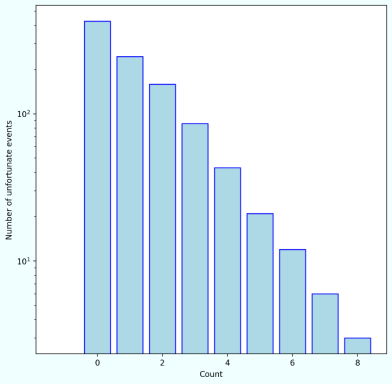}
\includegraphics[scale=0.4]{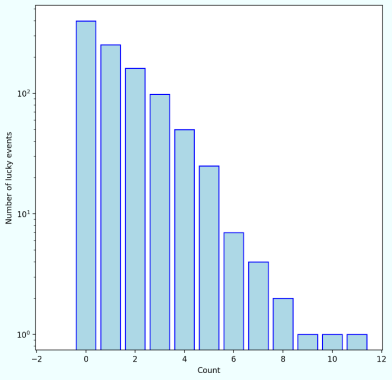}
\caption{The frequency distributions of numbers of unlucky and lucky events (Location-Random)}
\label{fig:lr12}
\end{figure}

\begin{figure}
\centering
\includegraphics[scale=0.4]{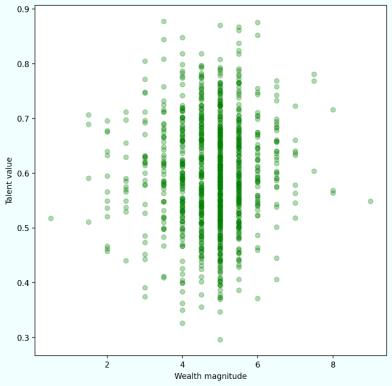}
\caption{Talent-Wealth plots(Location-Random)}
\label{fig:lr34}
\end{figure}

\begin{figure}
\centering
\includegraphics[scale=0.4]{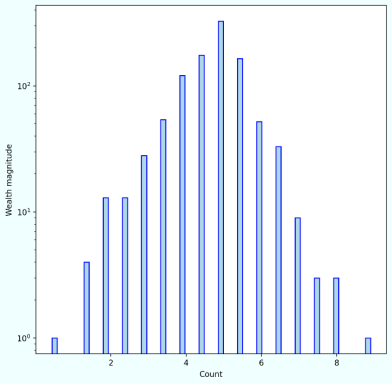}
\caption{The final distribution of wealth among the population (Location-Random)}
\label{fig:lr5}
\end{figure}

\begin{figure}
\centering
\includegraphics[scale=0.4]{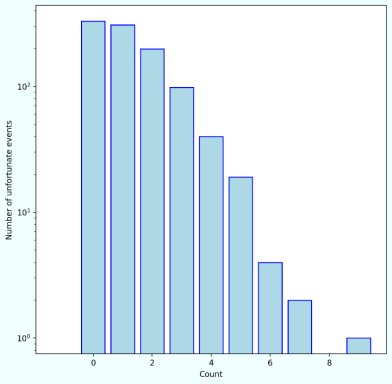}
\includegraphics[scale=0.4]{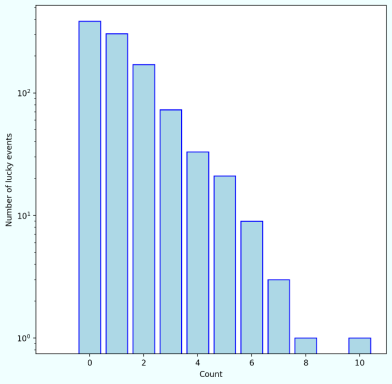}
\caption{The frequency distributions of numbers of unlucky and lucky events (Location-Highest)}
\label{fig:lh12}
\end{figure}

\begin{figure}
\centering
\includegraphics[scale=0.4]{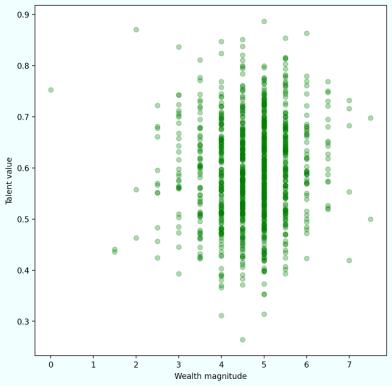}
\caption{Talent-Wealth plots(Location-Highest)}
\label{fig:lh34}
\end{figure}

\begin{figure}
\centering
\includegraphics[scale=0.4]{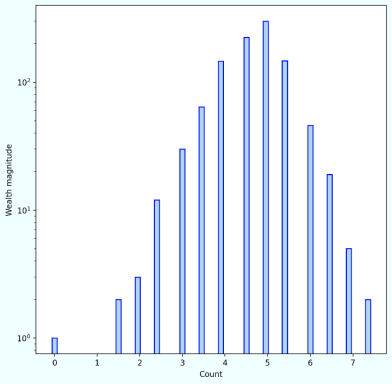}
\caption{The final distribution of wealth among the population (Location-Highest)}
\label{fig:lh5}
\end{figure}

\begin{figure}
\centering
\includegraphics[scale=0.4]{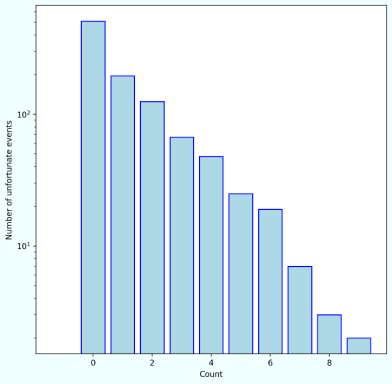}
\includegraphics[scale=0.4]{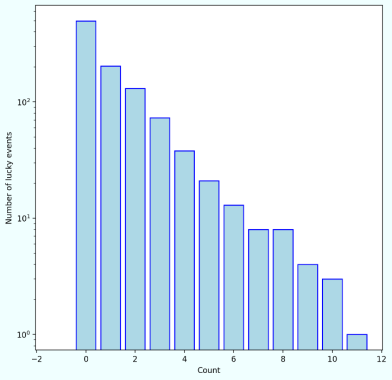}
\caption{The frequency distributions of numbers of unlucky and lucky events (Location-Average)}
\label{fig:la12}
\end{figure}

\begin{figure}
\centering
\includegraphics[scale=0.4]{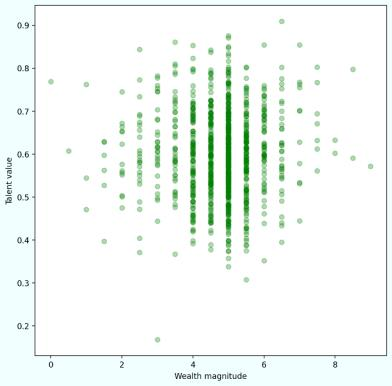}
\caption{Talent-Wealth plots(Location-Average)}
\label{fig:la34}
\end{figure}

\begin{figure}
\centering
\includegraphics[scale=0.4]{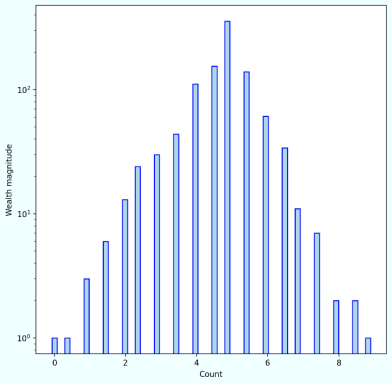}
\caption{The final distribution of wealth among the population (Location-Average)}
\label{fig:la5}
\end{figure}

\begin{figure}
\centering
\includegraphics[scale=0.4]{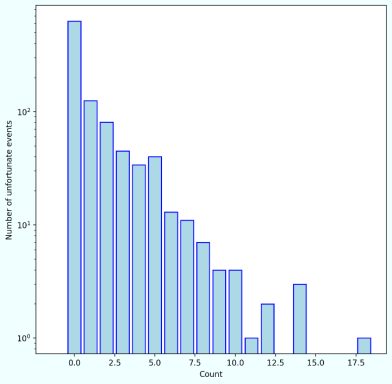}
\includegraphics[scale=0.4]{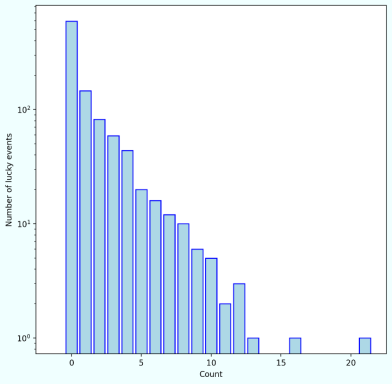}
\caption{The frequency distributions of numbers of unlucky and lucky events (Wealth-Random)}
\label{fig:wr12}
\end{figure}

\begin{figure}
\centering
\includegraphics[scale=0.4]{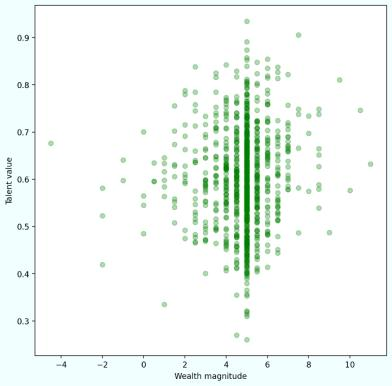}
\caption{Talent-Wealth plots(Wealth-Random)}
\label{fig:wr34}
\end{figure}

\begin{figure}
\centering
\includegraphics[scale=0.4]{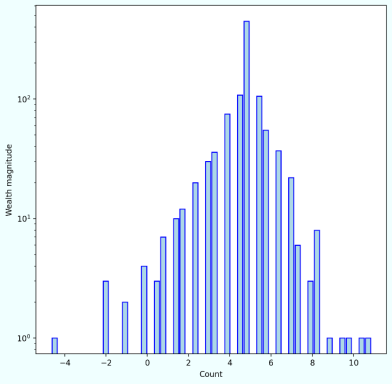}
\caption{The final distribution of wealth among the population (Wealth-Random)}
\label{fig:wr5}
\end{figure}

\begin{figure}
\centering
\includegraphics[scale=0.4]{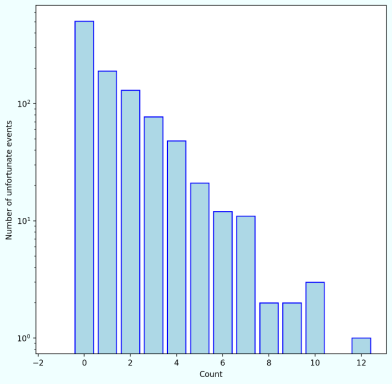}
\includegraphics[scale=0.4]{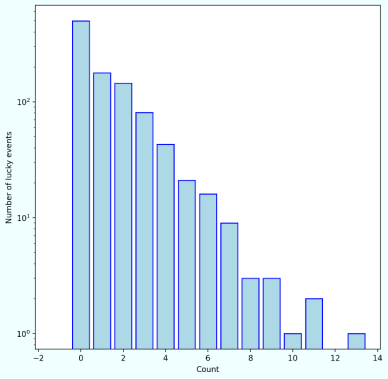}
\caption{The frequency distributions of numbers of unlucky and lucky events (Wealth-Highest)}
\label{fig:wh12}
\end{figure}

\begin{figure}
\centering
\includegraphics[scale=0.4]{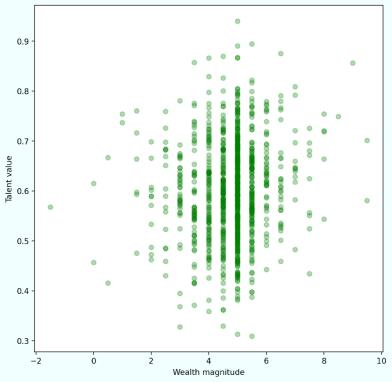}
\caption{Talent-Wealth plots(Wealth-Highest)}
\label{fig:wh34}
\end{figure}

\begin{figure}
\centering
\includegraphics[scale=0.4]{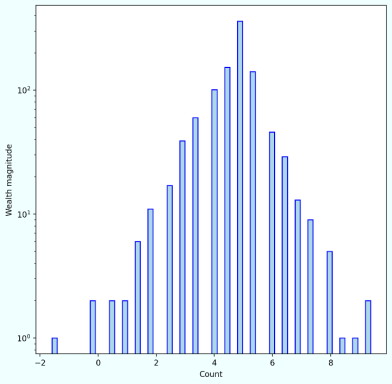}
\caption{The final distribution of wealth among the population (Wealth-Highest)}
\label{fig:wh5}
\end{figure}

\begin{figure}
\centering
\includegraphics[scale=0.4]{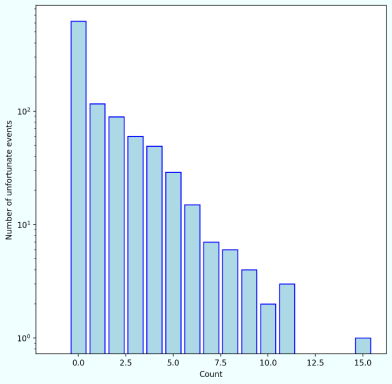}
\includegraphics[scale=0.4]{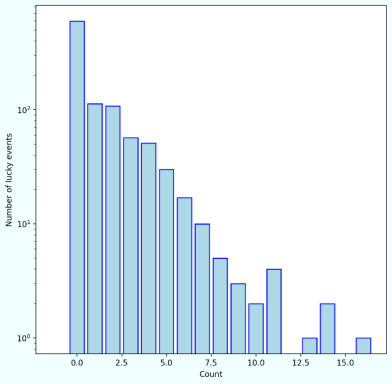}
\caption{The frequency distributions of numbers of unlucky and lucky events (Wealth-Average)}
\label{fig:wa12}
\end{figure}

\begin{figure}
\centering
\includegraphics[scale=0.4]{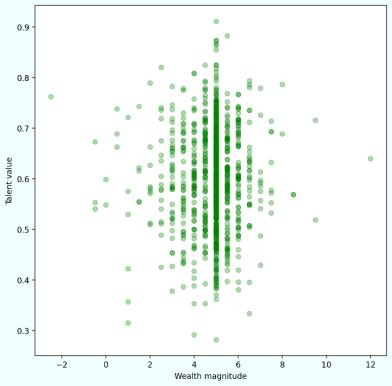}
\caption{Talent-Wealth plots(Wealth-Average)}
\label{fig:wa34}
\end{figure}

\begin{figure}
\centering
\includegraphics[scale=0.4]{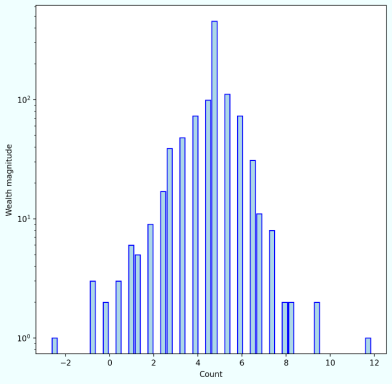}
\caption{The final distribution of wealth among the population (Wealth-Average)}
\label{fig:wa5}
\end{figure}

\begin{figure}
\centering
\includegraphics[scale=0.4]{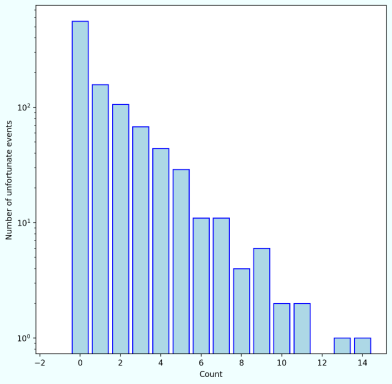}
\includegraphics[scale=0.4]{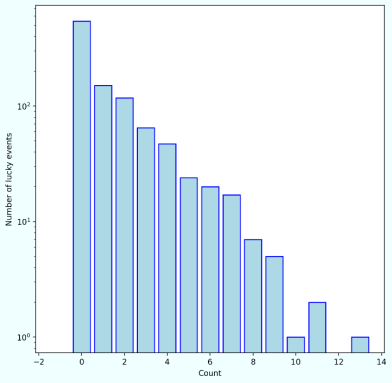}
\caption{The frequency distributions of numbers of unlucky and lucky events (Talent-Random)}
\label{fig:tr12}
\end{figure}

\begin{figure}
\centering
\includegraphics[scale=0.4]{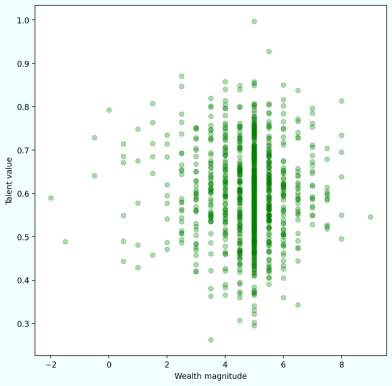}
\caption{Talent-Wealth plots(Talent-Random)}
\label{fig:tr34}
\end{figure}

\begin{figure}
\centering
\includegraphics[scale=0.4]{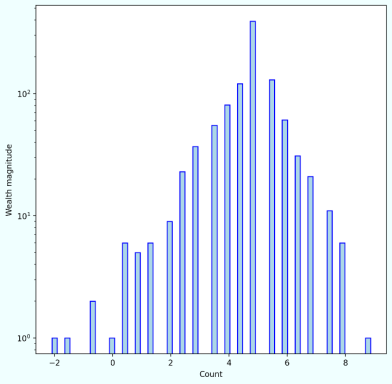}
\caption{The final distribution of wealth among the population (Talent-Random)}
\label{fig:tr5}
\end{figure}

\begin{figure}
\centering
\includegraphics[scale=0.4]{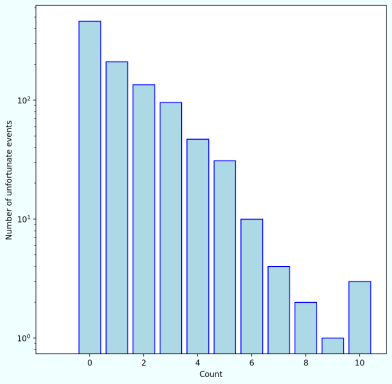}
\includegraphics[scale=0.4]{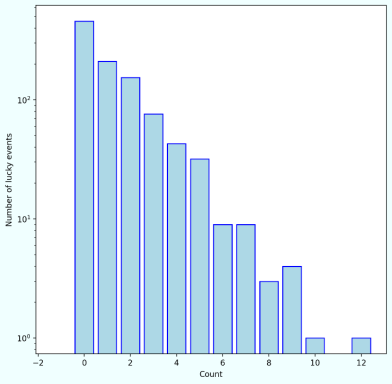}
\caption{The frequency distributions of numbers of unlucky and lucky events (Talent-Highest)}
\label{fig:th12}
\end{figure}

\begin{figure}
\centering
\includegraphics[scale=0.4]{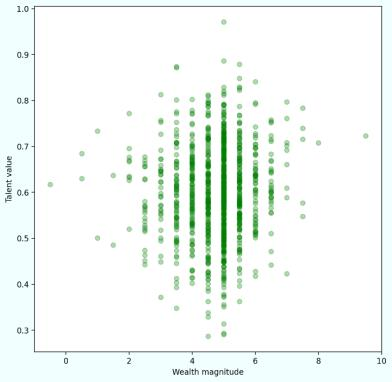}
\caption{Talent-Wealth plots(Talent-Highest)}
\label{fig:th34}
\end{figure}

\begin{figure}
\centering
\includegraphics[scale=0.4]{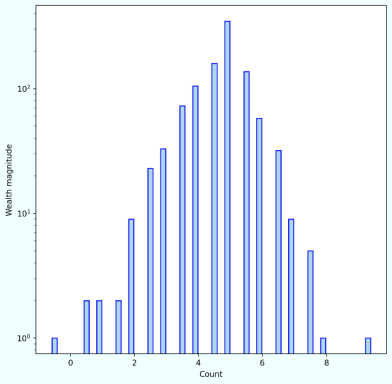}
\caption{The final distribution of wealth among the population (Talent-Highest)}
\label{fig:th5}
\end{figure}

\begin{figure}
\centering
\includegraphics[scale=0.4]{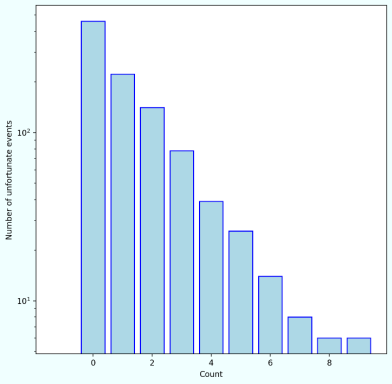}
\includegraphics[scale=0.4]{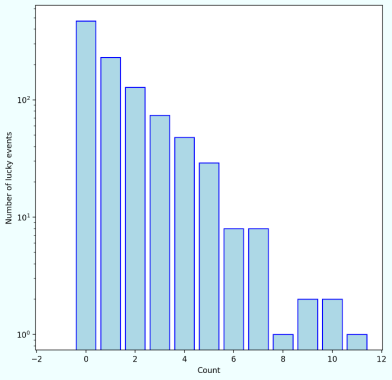}
\caption{The frequency distributions of numbers of unlucky and lucky events (Talent-Average)}
\label{fig:ta12}
\end{figure}

\begin{figure}
\centering
\includegraphics[scale=0.4]{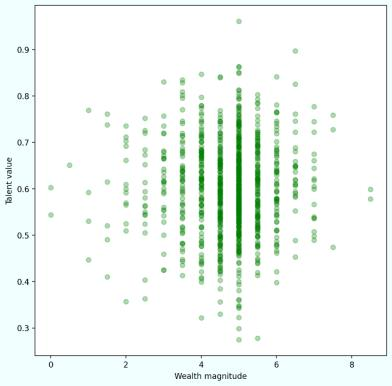}
\caption{Talent-Wealth plots(Talent-Average)}
\label{fig:ta34}
\end{figure}

\begin{figure}
\centering
\includegraphics[scale=0.4]{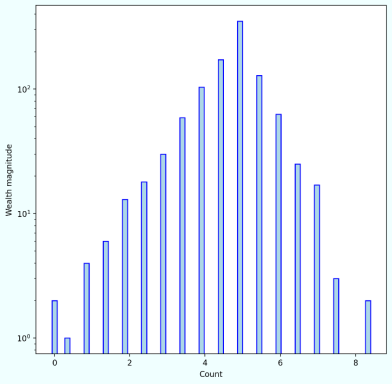}
\caption{The final distribution of wealth among the population (Talent-Average)}
\label{fig:ta5}
\end{figure}

We set population $N = 1000$, number of events $N_E = 500$, percent ratio of lucky events $pL = 50$, mean of talents $m_T = 0.6$, deviation of talents $\sigma_T = 0.1$, radius of events $r = 1$, moving speed of events $v = 1$, initial magnitude of wealth $C (0) = 5$, altitude of wealth magnitude changing $dC = 0.5$, radius of location neighborhood $R = 5$, radius of wealth neighborhood $nC = 3$, radius of talent neighborhood $nT = 1$ in our simulations. Fig.\ref{fig:rr12} to Fig.\ref{fig:rr5} show the simulation results with Random social network and Random mobile strategy. Fig.\ref{fig:rh12} to Fig.\ref{fig:rh5} show the simulation results with Random social network and Highest mobile strategy. Fig.\ref{fig:ra12} to Fig.\ref{fig:ra5} show the simulation results with Random social network and Average mobile strategy. Fig.\ref{fig:lr12} to Fig.\ref{fig:lr5} show the simulation results with Location social network and Random mobile strategy. Fig.\ref{fig:lh12} to Fig.\ref{fig:lh5} show the simulation results with Location social network and Highest mobile strategy. Fig.\ref{fig:la12} to Fig.\ref{fig:la5} show the simulation results with Location social network and Average mobile strategy. Fig.\ref{fig:wr12} to Fig.\ref{fig:wr5} show the simulation results with Wealth social network and Random mobile strategy. Fig.\ref{fig:wh12} to Fig.\ref{fig:wh5} show the simulation results with Wealth social network and Highest mobile strategy. Fig.\ref{fig:wa12} to Fig.\ref{fig:wa5} show the simulation results with Wealth social network and Average mobile strategy. Fig.\ref{fig:tr12} to Fig.\ref{fig:tr5} show the simulation results with Talent social network and Random mobile strategy. Fig.\ref{fig:th12} to Fig.\ref{fig:th5} show the simulation results with Talent social network and Highest mobile strategy. Fig.\ref{fig:ta12} to Fig.\ref{fig:ta5} show the simulation results with Talent social network and Average mobile strategy. Table \ref{tab:policies} shows the mean of wealth magnitude, standard deviation of wealth magnitude and Gini coefficient under different combinations of social network and mobile strategies.  

The experimental data from the strategy combination revealed the impact of different strategies on the mean and standard deviation of wealth magnitude as well as the Gini coefficient. Looking at the mean of wealth magnitude, the wealth social network with random or average mobile strategy showed the highest mean wealth at 4.82, whereas the location social network and highest mobile strategy had a lower average wealth at 4.70. In terms of standard deviation, the random social network with highest mobile strategy had the highest standard deviation at 1.33, while the location social network with highest mobile strategy had the lowest at only 0.83. The random social network and random or highest mobile strategy also had the highest Gini coefficient at 0.13, while the location social network with highest mobile strategy had the lowest at 0.09. These data demonstrate that the random and wealth social network perform better in achieving high mean wealth, but tend to with high standard deviation and Gini coefficient. On the other hand, the location social network performs better in standard deviation and Gini coefficient, but limits the increase in mean wealth. The result of talent social network is somewhere in between. Also, highest mobile strategy tends to suppress the mean and standard deviation of wealth and the Gini coefficient at the same time, while average mobile strategy tends to increase the mean of wealth magnitude while increasing the standard deviation and Gini coefficient except for random social network scenarios.  

\begin{table}[htbp]
\centering
\caption{Mean of wealth magnitude, standard deviation of wealth magnitude and Gini coefficient under
different combinations of social network and mobile strategies.}
\label{tab:policies}
\begin{tabular}{|l|c|c|c|}
\hline
Policies & Mean of wealth & Standard deviation of wealth & Gini coefficient \\
\hline
Random-Random & {\bf4.82} & 1.24 & 0.13 \\
Random-Highest & 4.75 & 1.33 & 0.13 \\
Random-Average & 4.75 & 1.10 & 0.12 \\
Location-Random & 4.79 & 0.94 & 0.10 \\
Location-Highest & 4.70 & {\bf0.83} & {\bf0.09} \\
Location-Average & 4.78 & 1.04 & 0.11 \\
Wealth-Random & {\bf4.82} & 1.32 & 0.13 \\
Wealth-Highest & 4.76 & 1.09 & 0.12 \\
Wealth-Average & {\bf4.82} & 1.15 & 0.12 \\
Talent-Random & 4.78 & 1.18 & 0.12 \\
Talent-Highest & 4.74 & 0.99 & 0.11 \\
Talent-Average & 4.74 & 1.02 & 0.11 \\
\hline
\end{tabular}
\end{table}

\section{Conclusion}

In this report, we focused on investigating the influence of various information exchange strategies on wealth distribution. We incorporated the effects of non-decouplable social networks and information exchange strategies into simulations of talent and luck, and analyze the effects of individual and group information exchange strategies and cultures on wealth distribution. As a preliminary research, we restricted the update rules of social networks to 4 types: random, location, wealth, and talent, and limit the mobile strategies to 3 types: random, highest, and average. We used 3 indicators, namely, mean wealth magnitude, standard deviation of wealth magnitude, and Gini coefficient to evaluate the macroscopic results of the simulation. The qualitative results of the research suggested that different social networks and information exchange strategies have varying impacts on wealth distribution. Specifically, the research found that the location social network tent to perform better in terms of standard deviation and Gini coefficient, but limits the increase in mean wealth. In contrast, the wealth social network performed better in achieving high mean wealth, but tent to have higher standard deviation and Gini coefficient. The talent social network falled somewhere in between. The choice of mobile strategy also had a significant impact on wealth distribution, with the highest mobile strategy suppressing both the mean and standard deviation of wealth and the Gini coefficient, while the average mobile strategy tent to increase the mean of wealth magnitude but also increase the standard deviation and Gini coefficient.

\bibliographystyle{plainnat}
\bibliography{main}












\end{document}